\pdfoutput=1

\documentclass[11pt]{article}

\usepackage[final]{acl}

\usepackage{times}
\usepackage{latexsym}

\usepackage[T1]{fontenc}

\usepackage[utf8]{inputenc}

\usepackage{microtype}

\usepackage{inconsolata}

\usepackage{hyperref}
\usepackage{url}
\usepackage{booktabs}
\usepackage{multirow}
\usepackage{multicol}
\usepackage{makecell}
\usepackage{graphicx}
\usepackage{wrapfig}
\usepackage{amsthm}
\usepackage{xcolor,colortbl}
\usepackage{amssymb}
\usepackage{pifont}
\usepackage{fdsymbol}
\usepackage{color, colortbl}
\usepackage{listings}
\usepackage{xspace}
\usepackage{subcaption}
\usepackage{mdframed}

\newcommand{\ds}{\textsc{ECCO}\xspace}

\usepackage{color-edits}
\addauthor{todo}{red}
\addauthor{zw}{orange}
\addauthor{df}{cyan}
\addauthor{vish}{blue}
\addauthor{sid}{red}

\title{ECCO: Can We Improve Model-Generated Code Efficiency\\Without Sacrificing Functional Correctness?}

\author{
Siddhant Waghjale\thanks{\ Equal contribution, ordered by a coin toss.} \hspace{1em} Vishruth Veerendranath\footnotemark[1] \hspace{1em} Zora Zhiruo Wang \hspace{1em} Daniel Fried \\
Language Technologies Institute, Carnegie Mellon University \\
\texttt{\{swaghjal, vveerend, zhiruow, dfried\}@cs.cmu.edu}
}

\begin{document}
\maketitle
\begin{abstract}
Although large language models (LLMs) have been largely successful in generating functionally correct programs, conditioning models to produce efficient solutions while ensuring correctness remains a challenge. Further, unreliability in benchmarking code efficiency is a hurdle across varying hardware specifications for popular interpreted languages such as Python. In this paper, we present \ds, a reproducible benchmark for evaluating program efficiency via two paradigms: natural language (NL) based code generation and history-based code editing. On \ds, we adapt and thoroughly investigate the three most promising existing LLM-based approaches: in-context learning, iterative refinement with execution or NL feedback, and fine-tuning conditioned on execution and editing history.
While most methods degrade functional correctness and moderately increase program efficiency, we find that adding execution information often helps maintain functional correctness, and NL feedback enhances more on efficiency. We release our benchmark to support future work on LLM-based generation of efficient code.\footnote{\url{https://github.com/CodeEff/ECCO}}
\end{abstract}

\section{Introduction}
The ability to write efficient code is a cornerstone of software development \citep{li2022competition}. While large language models (LLMs) have shown remarkable progress in generating functionally correct code \citep{roziere2023code,guo2024deepseek}, the ability to generate solutions that are both correct and efficient remains elusive \citep{shypula2021learning,shypula2024learning}. 

\begin{figure*}[t!]
\vspace{-2mm}
    \centering
    \includegraphics[width=\textwidth]{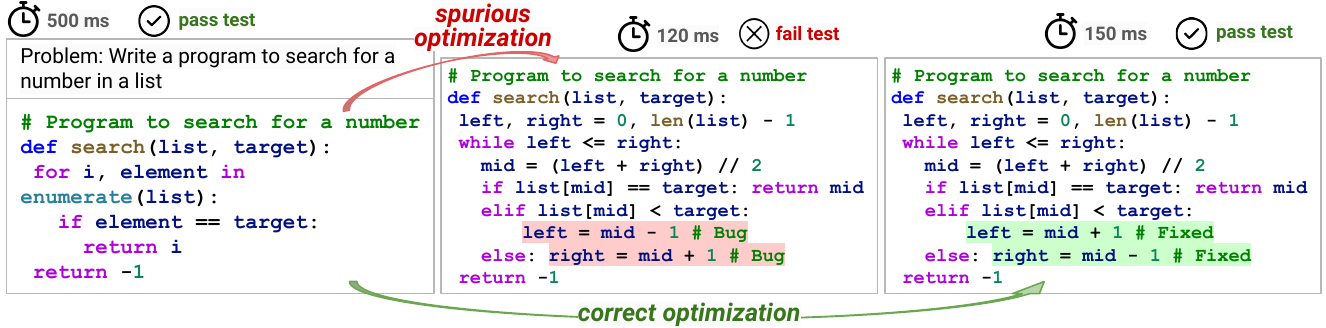}
    \caption{{\it Correctness-preserving} versus {\it spurious optimization} when optimizing a linear search algorithm with binary search on a sorted list. Spurious optimization can reduce runtime, but add errors that cause the program to be incorrect. In contrast, a true optimization reduces runtime and remains correct, as we emphasized in \ds.}
\label{fig:teaser}
\vspace{-2mm}
\end{figure*}

Current methods for optimizing program efficiency improve performance measured by execution time. However, this apparent success often comes at the cost of severely decreasing the functional correctness \citep{shypula2024learning}. An example of this issue is illustrated in \autoref{fig:teaser}: When optimizing the program on the left, models sometimes perform \textit{spurious optimizations} that, although they reduce the runtime, make the program no longer functionally correct so that it fails all test cases. On the other hand, a correct optimization (bottom right) --- that improves efficiency while maintaining functional correctness --- is often harder to achieve for current LMs. 
This \textit{spurious optimization} is certainly undesirable in practice, and can even increase debugging time for software developers \citep{li2022competition,cummins2023large}.
To achieve the goal of real and robust program optimization, we ask: \textit{Can LMs improve program efficiency without sacrificing functional correctness?}

In this work, we curate an efficiency-oriented programming benchmark \textbf{\ds}, short for \underline{E}nsuring \underline{C}orrectness in \underline{C}ode \underline{O}ptimizations, which enables program evaluation in three aspects: execution correctness, runtime efficiency, and memory efficiency. 
\ds supports two optimization paradigms: (i) history-based code editing: based on a previous version of the program, test if an LM can further optimize the code while maintaining its correctness, and (ii) NL-based code generation: test the efficiency of a program generated by an LM given a programming problem described in NL. We collect over 50$k$ Python solution pairs, spanning 1.3$k$ competitive programming problems \cite{puri2021codenet}, with an average of 3.1 public and 17.3 private test cases to support reliable execution-based evaluations of correctness and efficiency.

Further, to perform reliable and reproducible executions, we introduce an evaluation setup using a cloud-hosted code execution engine, \textsc{Judge0} \cite{dovsilovic2020judge0}, which produces stable execution output on correctness, runtime, and memory usage, thanks to its agnostic nature to local hardware specifications. It supports up to 66 programming languages (PLs), allowing future work to extend to other languages.

To explore various correctness-preserving program optimization methods, we evaluate three classes of methods on \ds --- in-context learning, iterative refinement, and fine-tuning, across a suite of open-source language models pre-trained on code --- which previous works have used to attempt to improve efficiency, while however overlooking their effects on correctness.
We find that execution information and fine-tuning help maintain functional correctness, and NL-involved prompting often yields higher efficiency improvements.
However, we broadly reconfirm findings that no existing methods can improve time/space efficiency without sacrificing functional correctness.
We hope \ds can serve as a solid testbed for program optimization, and call for more efforts in advancing correctness-preserving program optimizations.
\section{Related Work}

\paragraph{Benchmarks for Code Efficiency}
Some works have proposed benchmarks for optimizing program assembly code~\citep{bunel2016superoptimprog, shypula2021learning, shi2020symbsuperoptim, cummins2023metaoptim}. More recently, \citet{shypula2024learning} target C++ program speedups, and \citet{huang2024effibench} evaluate the efficiency of Python solutions for LeetCode coding interview problems~\citep{niu2024evaluating}.
Although most efforts on LLM-based code generation focus on evaluating functional correctness (e.g., \citealt{chen2021codexhumaneval}), some works evaluate code efficiency \citep{moudgalya2023tasty,sikka2020corcod,jeon2023deep,baik2024codecomplex} by classifying the time complexity of programs. 
However, these works are limited in their single-reference evaluation paradigm, assembly language support, or by the limited problem space on LeetCode~\citep{coignion2024performance}.
Our work supports reliable evaluation across arbitrary coding problems and the widely-used Python language.

\paragraph{Evaluating Program Efficiency}
It is challenging to robustly evaluate program efficiency, due to varying hardware platforms and setups.
Previous works have evaluated the efficiency of code by executing code in a local software environment \cite{singhal2024nofuneval, huang2024effibench} or using containerized environments on local hardware \citep{khan2023xcodeeval}, but this can result in varying runtime and memory usage across hardware, thus causing irreproducible evaluations. 
An alternative approach is to use an architecture simulator~\cite{shypula2024learning} which ensures the execution of each program is exactly simulated at the hardware level, but is limited to compiled languages such as C++. A hardware counter though reliable \cite{liu2024evaluating}, cannot measure memory usage. For popular interpreted languages such as Python and Java, some use LeetCode's execution engine~\citep{niu2024evaluating, coignion2024performance}, but with a restricted space of testable problems. In our work, we propose an evaluation setup using an accessible cloud computing instance that ensures consistent virtual hardware and reliable benchmarking.

\paragraph{Program Optimization Approaches}
To start, some works explore in-context learning to optimize program efficiency \citep{huang2024effibench}, with retrieval methods to select relevant examples \citep{gao2024search, shypula2024learning}.
Beyond vanilla prompting, iterative prompting methods \citep{madaan2024self, shinn2024reflexion, ridnik2024alphacodium} have been explored to improve specific aspects of generation, by incorporating feedback from an LM or external modules.
Meanwhile, finetuning has been proposed with self-play, synthetic preferences and problem-oriented data \cite{shypula2024learning, gee2024code, ye2024iterative}.
However, none of these methods have been rigorously studied for correctness-preserving optimization. We fill in this gap and provide systematic studies of all methods.

\section{The \ds~Benchmark}
\label{sec:3:benchmark}

In this section, we first introduce our evaluation platform (\S\ref{sec:3.1:eval-platform}), then describe the construction process of our \ds benchmark (\S\ref{sub:3.2:benchmark-construction}), and lastly, introduce our two task formulations with corresponding evaluation metrics (\S\ref{sub:3.3:task-formulation}).

\subsection{Evaluation Platform}
\label{sec:3.1:eval-platform}

To reliably evaluate program efficiency in both runtime and memory usage, we need to first establish a robust and reproducible evaluation platform.
However, evaluating program efficiency is challenging, as resource usage statistics vary greatly across hardware and setups \citep{singhal2024nofuneval, huang2024effibench}. 

\begin{figure}[ht]
    \centering
    \includegraphics[width=\linewidth]{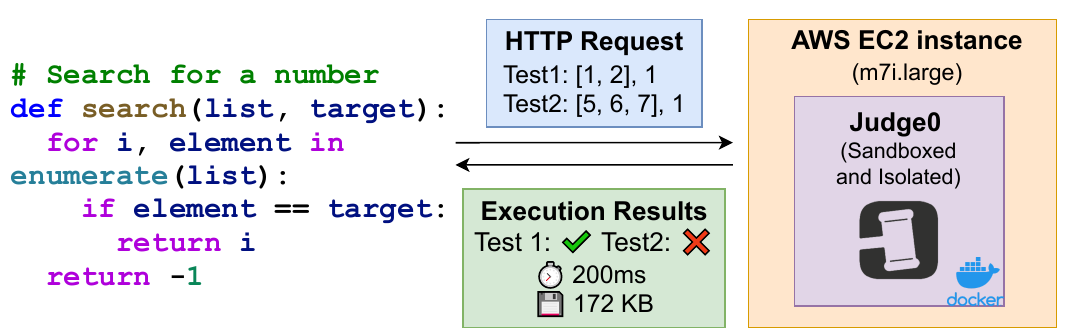}
    \caption{Evaluation platform using \textsc{Judge0}.}
    \label{fig:judge-setup}
    \vspace{-1mm}
\end{figure}

To ensure stable execution for interpreted languages such as Python, we propose to use a reproducible cloud compute instance that ensures the same virtual hardware, as illustrated in \autoref{fig:judge-setup}. 
Specifically, we use an EC2 instance (detailed in \S\ref{app:implementation}), and execute the code within a code execution engine \textsc{Judge0} \citep{dovsilovic2020judge0}. Note that our recipe can easily extend to over 60 programming languages that the \textsc{Judge0} engine supports. 
This is set up as a sandboxed Docker container within the instance, which thus ensures an isolated setup for secure and reproducible code execution. Our platform is similar to evaluating generated code on LeetCode execution console \cite{niu2024evaluating, coignion2024performance}, but applies to arbitrary coding problems and is not limited to questions available on LeetCode.\footnote{We make the Amazon Machine Image (AMI) for the setup available to enable reproducible benchmarks. We leave evaluations in other languages for future work.}

\subsection{Benchmark Construction} 
\label{sub:3.2:benchmark-construction}

Our goal is to collect programming problems, each with an NL description, and multiple functionally-correct solutions at varied efficiency levels.

\paragraph{Problem Selection}
We collect programming problems from the IBM CodeNet dataset \citep{puri2021codenet}, which contains competitive programming problems with NL descriptions, user program submissions, and other metadata, scraped from the AIZU and
AtCoder online judging systems. CodeNet problems mostly require algorithmic techniques such as data structure optimization.

Specifically, we first convert CodeNet to $\sim$187$k$ (slow, fast) Python code pairs following \citet{shypula2024learning}, where each pair of programs has two solutions for the same coding problem. We converted all Python 2 solutions to Python 3 using lib2to3.\footnote{\url{https://docs.python.org/3/library/2to3.html}}
Lastly, we filter out spurious program pairs in which the `fast' code was in fact slower when evaluated on our setup.\footnote{Speed statistics reported in CodeNet may be inconsistent.}

Next, we split the pairs and group programs by their associated problem ID. We filter out all problems with less than two solutions to ensure that each NL problem description has multiple associated solutions, to enable program optimization based on code editing history.
We then remove the programs that are repetitive or cannot successfully execute due to syntax errors or test case failures. 

Our curated dataset was partitioned into three subsets: train, validation, and test, with each split consisting of a distinct set of problems. In the end, the process yields 1,380 unique problems in \ds in total.

\paragraph{Test Case Collection} 
To evaluate functional correctness, we require test cases.
We collect (i) the original test cases for each problem from CodeNet, and (ii) the additional tests from the AlphaCode project \cite{li2022competition}. 
Each test case contains the program inputs as well as expected outputs when executing canonical program solutions on these inputs.
With these two sets of test cases, we simulate a realistic coding setting where one can refer to (i) as the public test cases for debugging or other accuracy-improving purposes, $T_{public}$, and (ii) as private test cases to conduct final execution-based evaluations on the programs, $T_{private}$.

\subsection{Task Formulation and Evaluation}
\label{sub:3.3:task-formulation}

We propose two formulations for the program optimization task, namely \textit{NL-instructed generation} and \textit{history-based program editing}. In this section, we introduce the data we use for each formulation, and our evaluations of program correctness, runtime, and memory usage.

\subsubsection{History-Based Program Editing}
\label{sub:3.3.2:history-based editing}

Our first paradigm follows previous work on program optimization \cite{shypula2024learning}, where we facilitate  a history-based editing paradigm. Concretely, we give a previous, presumably slow, version of the solution program, $p_{in}$. We then prompt LMs to edit the code to generate a more efficient version $p_{out}$, denoted as $CodeLM(p_{in}) \rightarrow p_{out}$, where $p_{out}$ is expected to run faster than $p_{in}$.

\paragraph{Evaluating Speedup and Memory Reduction}
Using the (slow, fast) program pairs remaining after post-processing in \S\ref{sub:3.2:benchmark-construction}, we evaluate the relative speedup and memory reduction of the model-generated program against the input program on private test cases $T_{private}$.

We adopt the \textit{speedup} metric introduced by \citet{shypula2024learning}, which is formulated as:
\begin{equation}
    Speedup = \frac{\text{Runtime of $p_{in}$}}{\text{Runtime of $p_{out}$}}
\end{equation}

Similarly, to evaluate improvement in memory usage, we introduce a \textit{memory reduction} metric as:
\begin{equation}
    Memory\ Reduction = \frac{\text{Memory of $p_{in}$}}{\text{Memory of $p_{out}$}}
\end{equation}

\begin{figure}[ht]
\vspace{-2mm}
    \centering
    \includegraphics[width=0.8\linewidth]{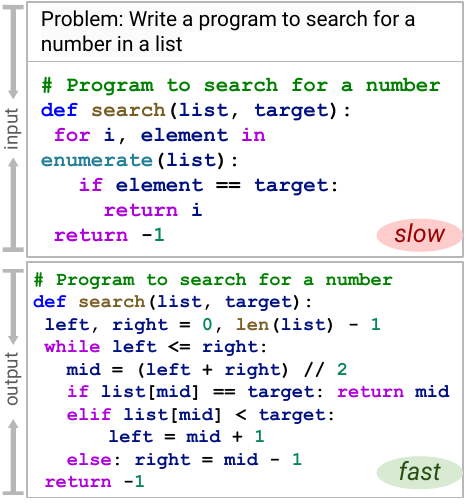}
    \caption{Illustration of history-based editing.}
    \label{fig:editing-example}
\end{figure}

\subsubsection{NL-Instructed Generation}
In addition, we support the most common NL-to-code generation setup: given the NL description $d$ of a problem, we ask the LM to generate the program solution $p$, as $CodeLM(d) \rightarrow p$.
Our goal is for the code LM to generate an efficient and correct solution $p$. We execute $p$ on the private test cases $T_{private}$ to evaluate its performance.

\begin{figure}[ht]
\vspace{-2mm}
    \centering
    \includegraphics[width=0.8\linewidth]{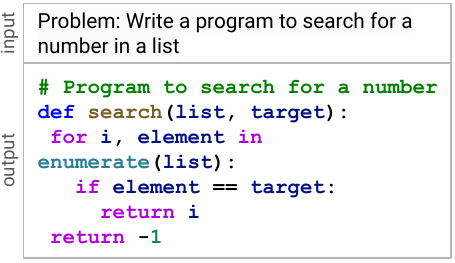}
    \caption{Illustration of NL-instructed generation.}
    \label{fig:nl2code-example}
\vspace{-2mm}
\end{figure}

\paragraph{Solution Program Spectrum}
To evaluate relative runtime and memory efficiency, we measure where a model-generated program lies on the spectrum of all user-submitted programs to that problem. 
We use the \textsc{Judge0} evaluation platform (\S\ref{sec:3.1:eval-platform}) to measure the runtime and memory usage.

\paragraph{Evaluating Percentile over the Spectrum}
We introduce \textit{runtime and memory percentile} to measure the efficiency of the model-generated program over the solution spectrum for a given problem as: 
\begin{equation}
    Runtime\ \% = \frac{\text{\# Slower user programs}}{\text{Total \# of user programs}}
\end{equation}
\begin{equation}
    Memory\ \% = \frac{\text{\# Programs w/ more memory}}{\text{Total \# of user programs}}
\end{equation}

\subsubsection{Evaluating Functional Correctness}
To measure if program correctness is preserved, a key metric is the functional correctness of model-generated programs. We adopt the pass@1 metric introduced by \citet{chen2021codexhumaneval}, which samples one program from the model and measures whether the generated program passes all test cases.

\subsection{\ds Feature Analysis}
\label{sub:3.3:feature-analysis}

After filtering the problem description dataset, we split the dataset into train, test, and validation sets for experiments. We ensure that no problem IDs overlap across these splits, to avoid data contamination. As shown by the detailed statistics of \ds in \autoref{tab:dataset-statistics}, \ds contains 1.3$k$ problems and over 50$k$ program pairs for code optimization evaluation.
\begin{table}[h]
\vspace{-2mm}
\centering
\resizebox{\linewidth}{!}{
\begin{tabular}{lcrrcc}
\toprule
\multirow{2}{*}{\bf Split} & \multirow{2}{*}{\bf \# Problems} & \multirow{2}{*}{\bf \# Pairs} & \multicolumn{2}{c}{\bf \# Avg. Test Cases}  \\
{} & {} & {} & {Public} & {Private} \\
\midrule
Train & 1262 & 48386  & 3.14 & {17.21} \\
Val & {~~~~69} & 2359 & 3.17 & {17.25} \\
Test & {~~~~48} & 794  & 3.29 & {20.00} \\
\bottomrule
\end{tabular}
}
\vspace{-1mm}
\caption{\ds dataset statistics.}
\label{tab:dataset-statistics}
\vspace{-4mm}
\end{table}

\begin{figure*}[t!]
\vspace{-2mm}
    \centering
    \includegraphics[width=\textwidth]{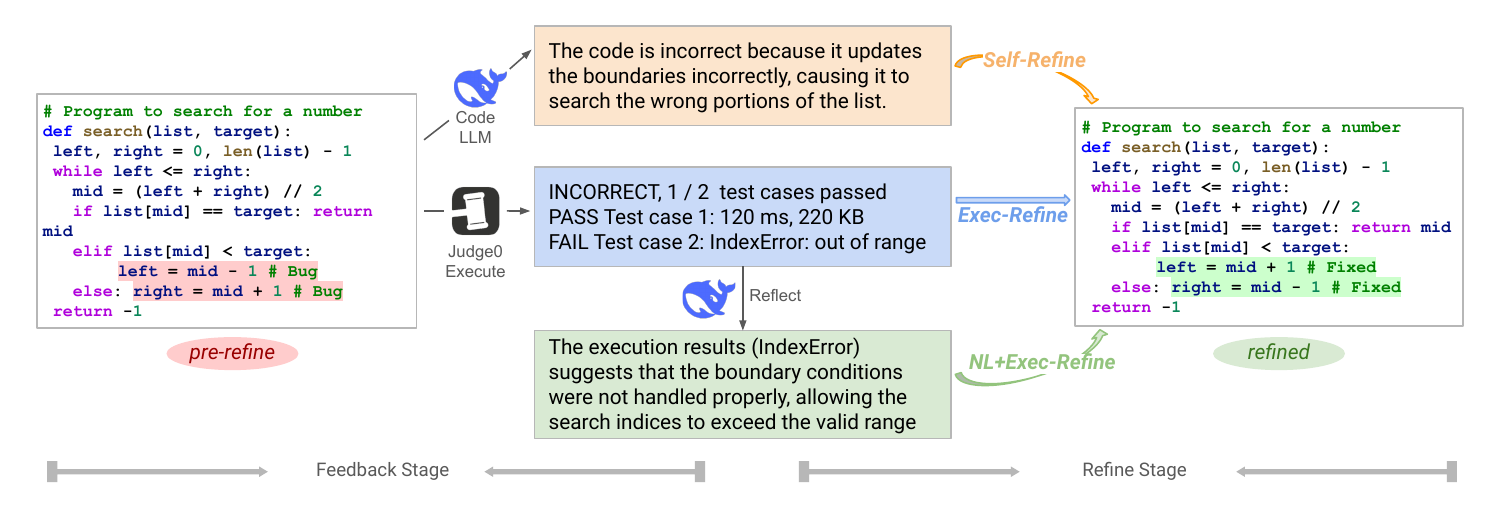}
    \caption{Iterative refinement methods utilizing different forms of feedback. \emph{Self-Refine} uses Natural Language feedback, \emph{Exec-Refine} uses raw execution results on $T_{public}$ and \emph{NL+Exec-Refine} uses NL reflection of execution.}
\label{fig:refinement-methods}
\vspace{-2mm}
\end{figure*}

\section{Efficiency-Improving Approaches}
\label{sec:5:approaches}

We explore various top-performing code generation approaches to improve program efficiency, while maintaining functional correctness, including in-context learning (\S\ref{sec:5.1:in-context-learning}), iterative refinement (\S\ref{sec:5.2:iterative-refinement}), and fine-tuning (\S\ref{sec:5.3:fine-tuning}).

\subsection{In-Context Learning}
\label{sec:5.1:in-context-learning}

We explore two mainstream prompting strategies: instruction prompting and few-shot learning. 

\paragraph{Instruction prompting}
Many LMs perform better when incorporating instructions \cite{ouyang2022training,wei2022finetuned}. We use two prompts: $I_{gen}$ for NL-based generation which instructs models to generate correct and efficient programs; and $I_{eff}$ for history-based editing which instructs models to optimize the input program. $I_{eff}$ is adapted from PIE \cite{shypula2024learning} and Self-Refine \cite{madaan2024self}. See \S\ref{app:prompt-details} for details.

\paragraph{Few-Shot Learning}
We add few-shot example demonstrations \citep{brown2020language}: for the NL-based setting, using (NL, fastest program) pairs; for history-based editing, using (slow program, fast program) pairs.
We randomly sample examples from the train set as the few-shot examples.

\subsection{Iterative Refinement}
\label{sec:5.2:iterative-refinement}
We explore three methods (illustrated in \autoref{fig:refinement-methods} ) to iteratively refine the generated code to be more efficient, which intuitively aligns with the way that humans improve code \citep{madaan2024self}.

\paragraph{Self-Refine with NL Feedback}
We adopt self-refine \citep{madaan2024self} that prompts LMs to iteratively examine the output and refine it. 
More concretely, (1) we first prompt the LM to generate a candidate solution; (2) we ask the same model to produce NL reasoning about why the code is incorrect and/or inefficient; and (3) we input the original input and the feedback from (2) to the model and ask it to generate an updated solution.

\paragraph{Exec-Refine with Interpreter Feedback}
We propose an alternative refinement strategy that obtains deterministic execution feedback from the interpreter, by running the program over $T_{public}$.\footnote{Models do not have access to the private test cases we finally evaluate on.}
If test cases are passed, the execution result provides the runtime and memory information; otherwise, this feedback provides interpreter error logs. Both correctness and efficiency can be informed via this feedbackare.

\paragraph{NL+Exec Refine: NL Feedback on Interpreter Results}
To allow feedback both in the forms of NL and execution outputs, we ground the LM feedback on execution results, inspired by the Reflexion feedback paradigm \cite{shinn2024reflexion}. Specifically, we first obtain the execution results as in \textit{exec-refine}, then ask the LM to write NL feedback on the incorrect/inefficient parts in the code, and use this as additional input in the refinement turn. 


\begin{table*}[ht]
\small
\centering
\resizebox{0.96\textwidth}{!}{
  \begin{tabular}{ll|ccc|ccc}
    \toprule
    \multicolumn{1}{c}{\multirow{2}{*}{\textbf{Model}}} & \multicolumn{1}{c|}{\multirow{2}{*}{\textbf{Setting}}} & \multicolumn{3}{c|}{\textbf{History-based Editing}} & \multicolumn{3}{c}{\textbf{NL-instructed Generation}} \\
    {} & {} & {pass@1} & {speedup} & {memory reduction} & {pass@1} & {runtime\%} & {memory\%} \\
    \midrule
    \multirow{2}{*}{StarCoder2} & {instruct}  & {49.4} & {1.49} & {1.24} & {~~4.2} & {50.64} & {55.72} \\
    {} & {few-shot} & {49.8} & {1.70} & {1.07} & {~~2.1} & {11.40} & {50.17} \\
    \midrule
    \multirow{2}{*}{CodeGemma} & {instruct}  & {42.5} & {1.43} & {1.10} & {18.8} & {41.70} & {51.83} \\
    {} & {few-shot} & {43.9} & {1.07} & {1.06} & {22.9} & {62.80} & {67.33} \\
    \midrule
    \multirow{2}{*}{WizardCoder} & {instruct} & {34.2} & {1.58} & {1.18} & {14.6} & {54.29} & {\bf84.53} \\
    {} & {few-shot}  & {27.4} & {1.38} & {1.12} & {14.6} & {58.69} & {71.00} \\
    \midrule
    \multirow{2}{*}{CodeLLaMa} & {instruct} & {57.5} & {1.44} & {1.11} & {~~8.3} & {45.30} & {74.18} \\
    {} & {few-shot} & {22.5} & {1.63} & {1.26} & {~~8.3} & {42.66} & {67.21} \\
    \midrule
    \multirow{2}{*}{DeepseekCoder} & {instruct} & {29.8} & {2.11} & {\bf 1.28} & {18.8} & {\bf59.01} & {75.86} \\
    {} & {few-shot}  & {35.2} & {\bf 2.26} & {1.20} & {22.9} & {55.52} & {66.09} \\
    \midrule
    \multirow{2}{*}{GPT-4o} & {instruct} & {\bf 66.6} & {1.64} & {1.10} & {\bf52.1} & {46.01} & {59.21} \\
    {} & {few-shot} & {65.8} & {1.62} & {1.12} & {41.7} & {49.87} & {64.44} \\
    \bottomrule
  \end{tabular}
}
\caption{Results using In-Context Learning approaches (instruction-prompting and few-shot learning)}
\vspace{-1em}
\label{tab:icl-results}
\end{table*}

\subsection{Fine-tuning}
\label{sec:5.3:fine-tuning}
We also explore three fine-tuning methods beyond prompting-alone approaches.

\paragraph{Vanilla Fine-tuning}
In this vanilla training setting, we leverage (NL, program) pairs and (slow program, fast program) pairs to train models independently for each paradigm. We format the data for both similarly to the in-context learning prompts (\S\ref{sec:5.1:in-context-learning}), and finetune on a causal language modelling task on the formatted data for each of the two paradigms independently.

\paragraph{Execution Conditioned Fine-tuning} 
Beyond fine-tuning with basic contexts, we posit that further conditioning on execution results could help. Therefore, we include execution results of PASS/FAIL status, runtime, and memory usage for each public test case for the input program. 


\paragraph{Trajectory Conditioned Fine-tuning}
For history-based editing, we further propose trajectory-conditioned fine-tuning, by adding a trajectory history of programs written by the same user for the given problem in context.
We first collect all problems with at least three programs submitted by the same user, and treat the series of programs as a trajectory. From each qualified trajectory, we designate the fastest code as the target output, and sample three other intermediate programs at the 0th, 33rd, and 66th percentile steps to use as inputs. We aim to allow the model to learn from the step-by-step improvements that led to the optimal solution, capturing the problem-solving process in addition to just the inputs and targets.
\section{Experiments}

\subsection{Experimental Setup}

\paragraph{Models}
We experiment with several best-performing LMs pre-trained on code. Specifically, we evaluate CodeLlama-13B \cite{roziere2023code}, DeepSeekCoder-7B-v1.5 \cite{guo2024deepseek}, CodeGemma-7B \cite{team2024gemma}, WizardCoder-13B-Python \cite{luo2023wizardcoder}, StarCoder2-15B \cite{lozhkov2024starcoder}. We use the instruction-tuned versions of all of these open-checkpoint models unless indicated otherwise. We also use the proprietary GPT-4o model for no-training methods.

\subsection{Results and Analysis}

\subsubsection{In-Context Learning}
As shown in \autoref{tab:icl-results}, all methods either reduce pass@1 of the program by a large margin (in editing mode) or obtain low pass@1 (generation mode).
Comparing the two paradigms, history-based editing results in a substantially higher pass@1 by referring to a base correct program, compared to NL-instructed generation which lacks a base program to start from. GPT-4o obtains a much higher pass@1 than all models in both paradigms.

\paragraph{History-based editing}
While in-context learning can effectively speed up the input program by $7$--$126$\%, but compromises correctness, dropping it to $22.5$--$66.6$, and uses more memory. \textit{Few-shot} shows this trend more explicitly than \textit{instruct}. Besides the limitations of LMs, this may be caused by the sampled few-shot demonstrations being algorithmically less relevant to the problem at hand.

\paragraph{NL-instructed generation}

While DeepseekCoder and CodeGemma's pass@1 improves by $4.1\%$ with \textit{few-shot}, GPT-4o and StarCoder2's pass@1 drops by $2.1-10.4\%$. Similarly for the efficiency metrics, CodeGemma and GPT-4o see an improvement whereas other models do not. GPT-4o significantly outperforms other models at pass@1 by 2.2$\times$, however there is no clear winner for efficiency. This highlights the complex trade-offs between correctness and efficiency specifically in the NL-instructed generation task.



\begin{table*}[ht]
  \vspace{-3mm}
    \centering
    \small
    \resizebox{0.95\textwidth}{!}{
    \begin{tabular}{ll|ccc|ccc}
      \toprule
      \multicolumn{1}{c}{\multirow{2}{*}{\textbf{Model}}} & \multicolumn{1}{c|}{\multirow{2}{*}{\textbf{Setting}}} & \multicolumn{3}{c|}{\textbf{History-based Editing}} & \multicolumn{3}{c}{\textbf{NL-instructed Generation}} \\
      {} & {} & {pass@1} & {speedup} & {memory reduction} & {pass@1} & {runtime\%} & {memory\%}\\
      \midrule
      \multirow{4}{*}{StarCoder2} & {pre-refine} & {49.4} & {1.49} & {1.24} & {~~4.2} & {50.64} & {55.72} \\
      \cmidrule{2-8}
      {} & {self-refine} & {26.7} & {1.55} & {\bf 1.30} & {~~2.1} & {~~5.79} & {71.50} \\
      {} & {exec-refine} & {\bf 39.5} & {1.49} & {1.23} & {~~2.1} & {\bf 29.27} & {55.79} \\
      {} & {nl+exec refine} & {26.1} & {\bf 2.13} & {1.26} & {~~2.1} & {5.79} & {\bf 81.69} \\
      \midrule
      \multirow{4}{*}{CodeGemma} & {pre-refine} & {42.5} & {1.43} & {1.10} & {18.8} & {41.70} & {51.83} \\
      \cmidrule{2-8}
      {} & {self-refine} & {15.1} & {\bf 2.08} & {\bf 1.15} & {~~6.3} & {\bf 41.23} & {\bf 59.18} \\
      {} & {exec-refine} & {\bf 33.2} & {1.59} & {1.12} & {\bf18.8} & {39.26} & {54.81} \\
      {} & {nl+exec refine} & {29.8} & {1.54} & {1.14} & {14.6} & {33.24} & {38.70} \\
      \midrule
      \multirow{4}{*}{WizardCoder} & {pre-refine} & {34.2} & {1.58} & {1.18} & {14.6} & {54.29} & {84.53} \\
      \cmidrule{2-8}
      {} & {self-refine} & {~~8.5} & {2.16} & {1.23} & {~~8.3} & {\bf 44.86} & {\bf 88.77} \\
      {} & {exec-refine} & {\bf 20.9} & {1.60} & {1.13} & {\bf12.5} & {44.12} & {76.86} \\
      {} & {nl+exec refine} & {18.3} & {\bf 2.90} & {\bf 1.30} & {12.5} & {31.92} & {79.19} \\
      \midrule
      \multirow{4}{*}{CodeLLaMa} & {pre-refine}  & {57.5} & {1.44} & {1.11} & {~~8.3} & {45.30} & {74.18} \\
      \cmidrule{2-8}
      {} & {self-refine} & {15.8} & {\bf 2.02} & {\bf 1.22} & {~~2.1} & {32.16} & {\bf 99.42} \\
      {} & {exec-refine} & {\bf 54.6} & {1.51} & {1.12} & {\bf~~4.2} & {44.09} & {85.28} \\
      {} & {nl+exec refine} & {16.2} & {1.37} & {1.02} & {\bf~~4.2} & {\bf 66.00} & {70.79} \\
      \midrule
      \multirow{4}{*}{DeepseekCoder} & {pre-refine} & {29.8} & {2.11} & {1.28} & {18.8} & {59.01} & {75.86} \\
      \cmidrule{2-8}
      {} & {self-refine} & {13.6} & {2.73} & {1.35} & {~~8.3} & {29.65} & {65.26} \\
      {} & {exec-refine} & {\bf 27.4} & {2.34} & {1.24} & {\bf20.8} & {\bf55.08} & {73.78} \\
      {} & {nl+exec refine} & {19.6} & {\bf 3.54} & {\bf 1.37} & {14.6} & {49.08} & {\bf85.15} \\
      \midrule
      \multirow{4}{*}{GPT-4o} & {pre-refine} & {66.6} & {1.64} & {1.10} & {52.1} & {46.01} & {59.21} \\
      \cmidrule{2-8}
      {} & {self-refine} & {47.8} & {\bf2.72} & {\bf1.25} & {37.5} & {51.12} & {44.03} \\
      {} & {exec-refine} & {\bf60.8} & {2.19} & {1.22} & {\bf52.1} & {\bf52.55} & {\bf59.47} \\
      {} & {nl+exec refine} & {58.8} & {2.39} & {1.22} & {47.9} & {49.79} & {47.46} \\
      \bottomrule
    \end{tabular}
    }
    \caption{Results with iterative refinement approaches. Feedback in NL (self \& nl+exec) improves efficiency better, whereas raw execution feedback (exec) maintains correctness more effectively.}
    \label{tab:refine-results}
  \vspace{-2mm}
  \end{table*}


\subsubsection{Iterative Refinement}
\autoref{tab:refine-results} shows all results with iterative refinement methods.
As a reference for the refinement approaches, we measure the LM-generated code in the first attempt at optimization without any refinement, and denote this method as \textit{pre-refine}. 

\paragraph{History-based editing paradigm} 
While all methods can effectively speed up the program, methods that involve NL feedback (\textit{self-refine} and \textit{nl+exec refine}) achieve the highest speedup across models. \textit{exec-refine} consistently yields the highest pass@1 for all models, by $3.4$--$38.8$ points more than the other two methods. 
We conjecture that execution outputs are better representations to inform functional correctness than NL descriptions.
Although it is easier to convey high-level optimization strategies in NL, conveying the functional correctness is harder. 
Overall, although the models are instructed to emphasize both correctness and efficiency, there seems to be an implicit trade-off between them. Additional analysis is in \S\ref{appendix:refinement_per_opt}.

\paragraph{NL-instructed generations}
We observe similar patterns as the editing mode, that \textit{exec-refine} best maintains functional correctness, and two other NL-involved approaches improve runtime/memory efficiency.
Compared to the in-context learning results in \autoref{tab:icl-results}, iterative refinement significantly improved memory\% for all the models, with the best method showing an average improvement of $12.06\%$ over the instruct method. However, the impact on runtime\% shows varying results among the different models.


\begin{figure*}[t]
\centering
    \begin{subfigure}{0.32\textwidth}
        \centering
        \includegraphics[width=\linewidth]{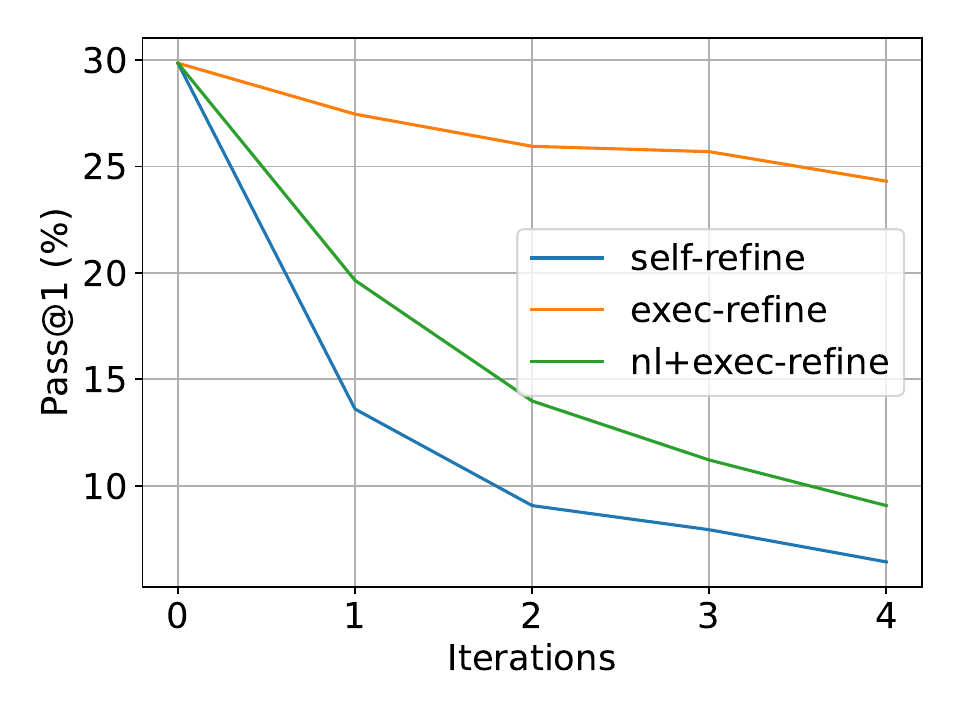}
        \caption{Pass@1}
        \label{fig:refine-pass}
    \end{subfigure}
    \begin{subfigure}{0.32\textwidth}
        \centering
        \includegraphics[width=\linewidth]{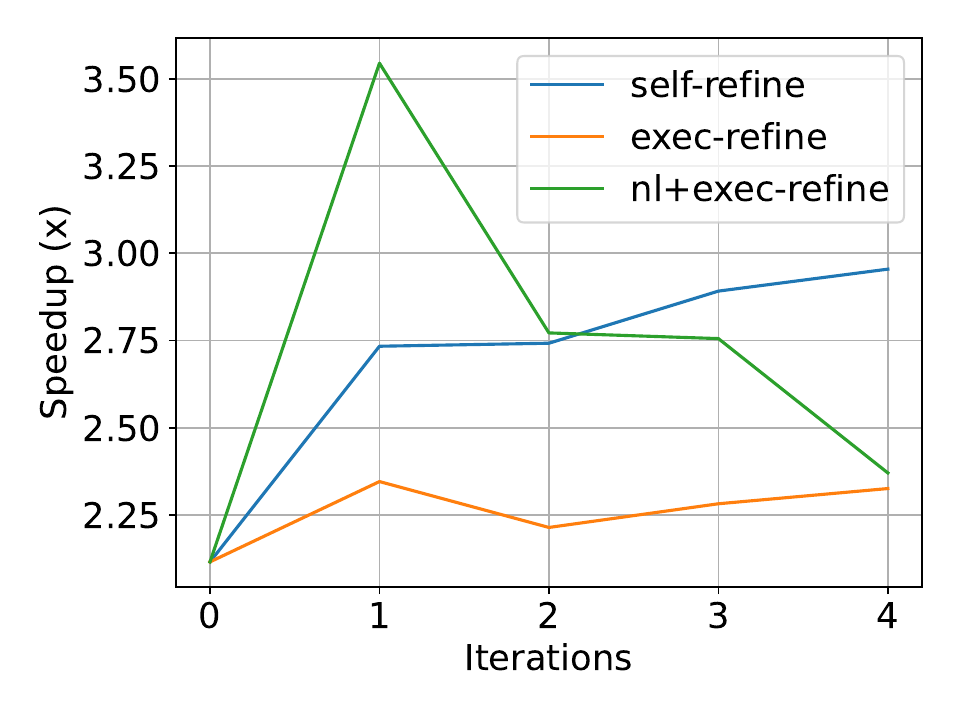}
        \caption{Speedup}
        \label{fig:refine-speedup}
    \end{subfigure}
    \begin{subfigure}{0.32\textwidth}
        \centering
        \includegraphics[width=\linewidth]{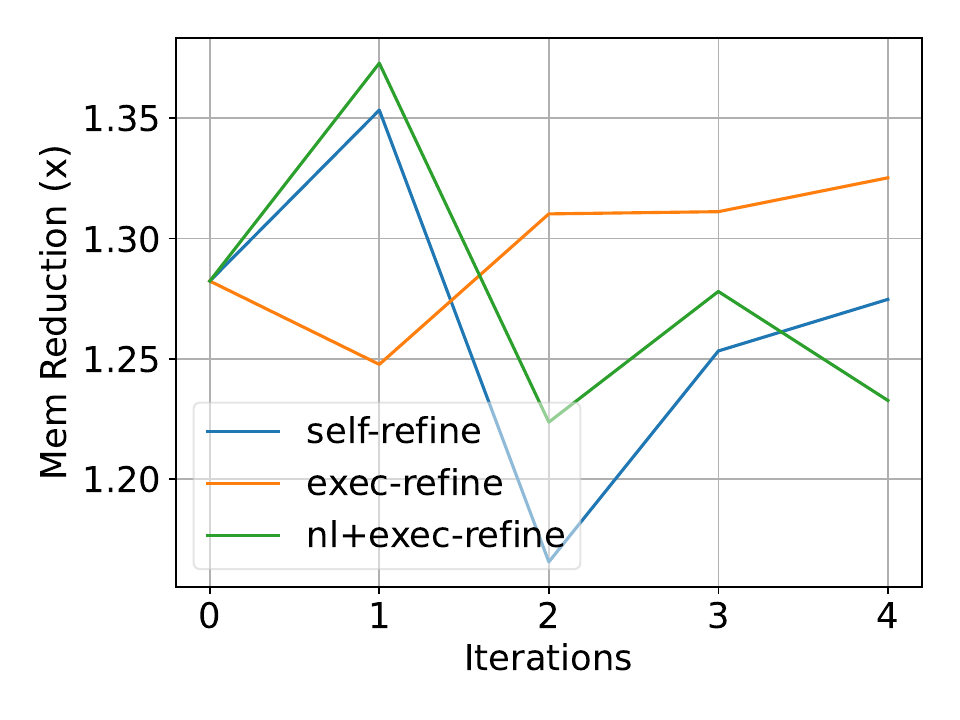}
        \caption{Memory Reduction}
        \label{fig:refine-memory}
    \end{subfigure}
\vspace{-1mm}
\caption{Performance of DeepseekCoder over multiple iterations of refinement. \emph{The improvement in efficiency is outweighed by the consistent drop in pass@1}. }
\label{fig:refine-results}
\end{figure*}

\subsubsection{Fine-tuning}

We perform parameter-efficient fine-tuning on CodeLLaMa-7B and DeepseekCoder-7B, the best-performing classes of models on the correctness and efficiency metrics in our prompting experiments respectively.

\paragraph{History-based editing}
As shown in \autoref{tab:editing-finetune}, fine-tuning is the most effective method in maintaining correctness in the editing paradigm. Especially for DeepseekCoder, compared to the highest prompting results $35.2$ using \textit{few-shot} examples, \textit{vanilla} and \textit{execution}-conditioned tuning improves by $6.9$ and $7.8$ points, and \textit{trajectory}-conditioned tuning further gains a $34.6$ point increase overall. This suggests that adding user-specific coding trajectories can help ground models into the optimization mode and substantially improve output correctness.

\begin{table}[h]
\centering
\resizebox{\linewidth}{!}{
    \begin{tabular}{ll|ccc}
    \toprule
    {\textbf{Model}} & {\textbf{Method}} & \textbf{pass@1} & \textbf{speedup} & \textbf{mem.red.} \\
    \midrule
    \multirow{3}{*}{CodeLLaMa-7B} & {Vanilla} & {43.0} & {1.11} & {1.01} \\
    {} & {Execution} & {45.0} & {\bf1.41} & {\bf1.04} \\
    {} & {Trajectory} & {\bf70.2} & {1.01} & {1.00} \\
    \midrule
    \multirow{3}{*}{DeepseekCoder} & {Vanilla} & {42.1} & {1.11} & {1.01} \\
    {} & {Execution} & {43.0} & {\bf 1.16} & {\bf 1.02} \\
    {} & {Trajectory} & {\bf 69.8} & {1.01} & {1.00} \\
    \bottomrule
    \end{tabular}
}
\caption{Fine-tuning results for history-based editing.}
\label{tab:editing-finetune}
\vspace{-2mm}
\end{table}

\paragraph{NL-instructed generation}
In the more complex NL-instructed generation task shown in \autoref{tab:nl2code-finetune}, fine-tuning is effective in improving the efficiency for CodeLLaMa but not for DeepseekCoder, highlighting the need for more robust fine-tuning methods that can handle trade-offs and effectively maintain correctness in this setting.

However, fine-tuning results in a much less efficiency improvement than prompting-based methods for both paradigms, possibly due to the limited power of parameter efficient fine-tuning.

\begin{table}[h]
\centering
\resizebox{\linewidth}{!}{
    \begin{tabular}{ll|ccc}
    \toprule
    {\textbf{Model}} & {\textbf{Method}} & \textbf{pass@1} & \textbf{runtime\%} & \textbf{mem\%} \\
    \midrule
    \multirow{2}{*}{CodeLLaMa-7B} & {Pre-trained} & {8.3} & {36.74} & {44.15}\\
     & {Finetuned} &        {8.3} & {\bf46.81} & {\bf71.32} \\
    \midrule
    \multirow{2}{*}{DeepseekCoder} & {Pre-trained} & {\bf18.8} & {\bf59.01} & {\bf75.86} \\
     & {Finetuned} & {16.7} & {43.61} & {67.31} \\
    \bottomrule
    \end{tabular}
}
\caption{Fine-tuning results for NL-based generation.}
\label{tab:nl2code-finetune}
\vspace{-2mm}
\end{table}
\section{Additional Analysis}

\paragraph{Does instruction tuning help in-context learning?}
As instruct model versions are expected to be better at in-context learning than their base counterparts, we compare base and instruct model versions in the editing paradigm.
In \autoref{tab:icl_basevinstruct}, the base versions get much higher correctness albeit at lower efficiency, showing that base and instruct versions lie at different points on the correctness-efficiency trade-off. As we emphasize both correctness and efficiency aspects in the NL instruction, we conjecture the instruct models take more hints from the input format and emphasize efficiency, while base models primarily emphasize correctness.


\begin{table}[ht]
\centering
\resizebox{\linewidth}{!}{
\begin{tabular}{ll|ccc}
\toprule
\multicolumn{1}{c}{\textbf{Model}} & \textbf{Version} & \textbf{pass@1} & \textbf{speedup} & \textbf{mem.red.}  \\
\midrule
\multirow{2}{*}{CodeLlama} & Instruct & {22.5} & \textbf{1.63} & \textbf{1.26} \\
 & Base & \textbf{46.4} & {1.02} & {1.00} \\
\midrule
\multirow{2}{*}{DeepseekCoder} & Instruct & {35.2} & \textbf{2.26} & \textbf{1.20} \\
 & Base & \textbf{45.4} & {1.04} & {1.00} \\
\midrule
\multirow{2}{*}{CodeGemma} & Instruct & {43.9} & \textbf{1.07} & \textbf{1.06} \\
 & Base & \textbf{48.2} & {1.01} & {1.00} \\    
\midrule
\multirow{2}{*}{StarCoder2} & Instruct & {49.8} & \textbf{1.70} & \textbf{1.07} \\
 & Base & \textbf{41.5} & {1.03} & {1.00} \\    
\bottomrule
\end{tabular}
}
\caption{Comparing base and instruct model versions with in-context learning methods.}
\label{tab:icl_basevinstruct}
\end{table}

\paragraph{Does multi-iteration refinement help?}

Multiple refinement iterations may improve results by allowing more turns for models to refine. To verify this, we evaluate the \textit{iterative refinement} method using 1--4 iterations. 
While self-refine and exec-refine improve program speedup over iterations (\autoref{fig:refine-speedup}), all methods continuously degrade the pass@1 to various extents. Exec-refine preserves correctness more effectively in further iterations as well.
For memory usage (\autoref{fig:refine-memory}), exec-refine consistently reduces memory usage, yet other methods exhibit big fluctuations.
In general, more iterations can speed up the program, yet further sacrifice correctness, thus has limited gains in terms of correctness-preserving optimization.

\begin{figure*}[t]
\vspace{-3mm}
\centering
\includegraphics[width=\linewidth]{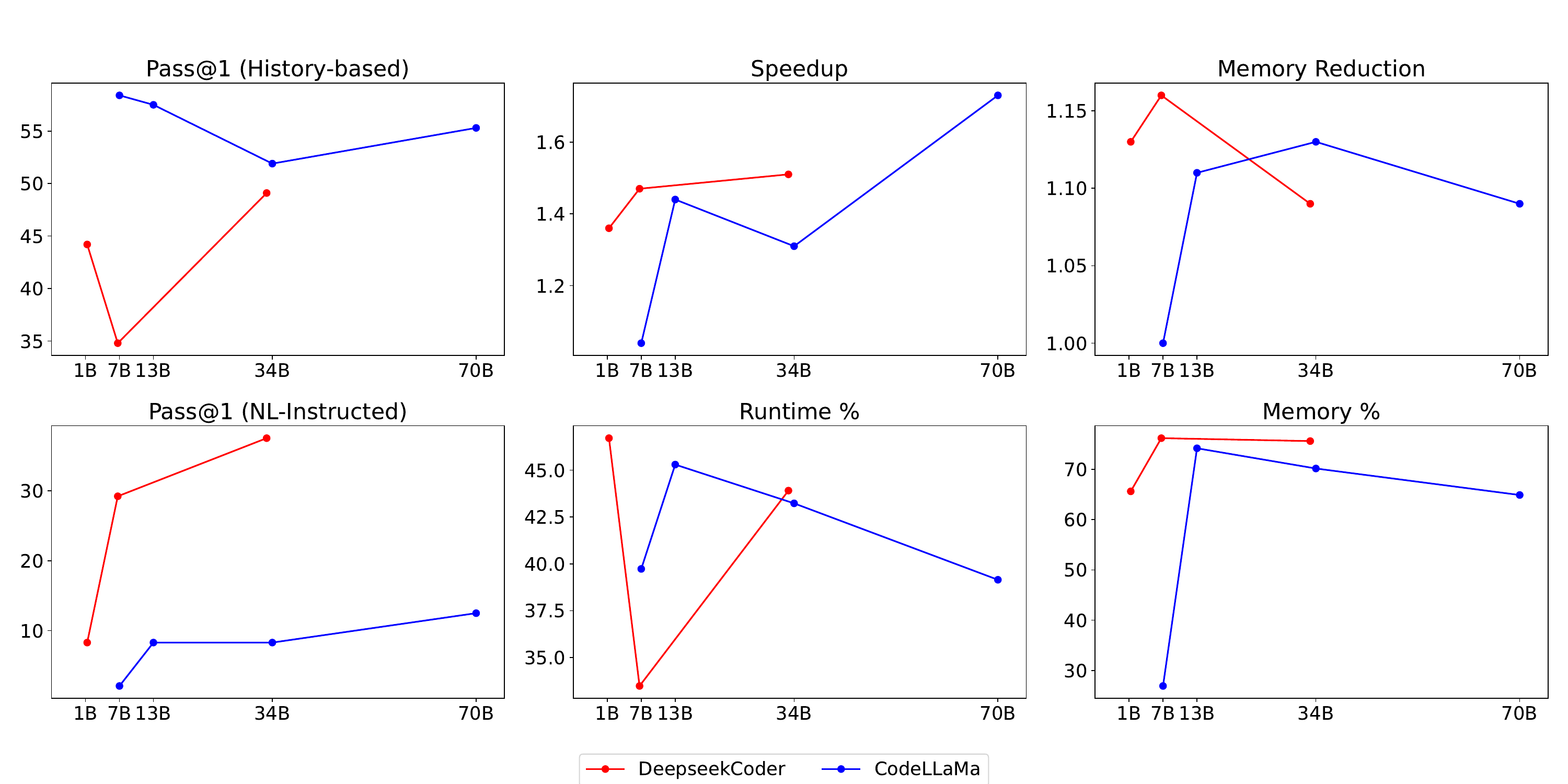}
\caption{Correctness and efficiency of generated code across model sizes for CodeLLaMa and DeepseekCoder. The top row corresponds to History-Based Editing and the bottom row includes NL-Instructed Generation.}
\label{fig:scaling}
\end{figure*}

\paragraph{Can iterative prompting fix incorrect solutions?}
To study whether models can recover from incorrect starting solutions in the history-based editing paradigm, we evaluate on a collection of 157 pairs of programs, \textsc{ECCO-Fix}, where the input code is almost correct: one that passed all public test cases but fails a few private test cases. 
As shown by \autoref{tab:ecco-spurious}, we show that \textit{exec-refine} can fix incorrect programs to pass all public and private test cases, with access to only the PASS/FAIL status of public test cases. In comparison, \textit{self-refine} breaks the correctness of more programs. 

Aligning with our findings in earlier sections and \S\ref{appendix:refinement_per_opt}, \textit{exec-refine}, with execution information in contexts, can encourage models to generate functionally correct programs.

\begin{table}[ht]
\centering
\resizebox{\linewidth}{!}{
\begin{tabular}{l|ccc}
  \toprule
  \multicolumn{1}{c|}{\bf Model} & {\bf Instruct} & {\bf Self-Refine} & {\bf Exec-Refine} \\
  \midrule
  StarCoder2 & 12.1 & \bf 9.5 & 8.9 \\
  CodeGemma & 10.8 & 3.1 & 10.8 \\
  WizardCoder & \bf 20.3 & 1.9 & 14.6 \\
  CodeLlama & 17.1  & 7.6 & \bf 36.9 \\
  DeepseekCoder & 12.7 & 5.7 & 15.2 \\
  \bottomrule
\end{tabular}
}
\caption{Pass@1 of iterative refinement strategies for models on \textsc{ECCO-Fix}, under the editing paradigm.}
\label{tab:ecco-spurious}
\end{table}

\paragraph{How does scale of model impact efficiency and correctness?}
We perform experiments across model scales for the CodeLLaMa (7B, 13B, 34B, 70B) and DeepseekCoder-v1 (1.3B, 6.7B, 33B) instruction-tuned model families. As seen in \autoref{fig:scaling} and \autoref{tab:scaling-results}, for both History-based Editing and NL-Instructed Generation, scaling model size presents a mix of benefits and trade-offs in terms of efficiency and correctness. For History-based Editing, larger CodeLLaMa models (such as 70B) show better speedup but experience diminishing gains in correctness (Pass@1), with a slight decrease in memory reduction. DeepseekCoder follows a similar pattern but consistently underperforms CodeLLaMa in correctness. In NL-Instructed Generation, scaling models significantly improves correctness, as seen with CodeLLaMa 70B and DeepseekCoder 33B, although  giving varying results for runtime and memory percentiles. 

\section{Conclusion}
In this paper, we introduce the \ds benchmark that enables two paradigms for Python program optimization, using \textsc{Judge0}, a language and platform-agnostic execution framework. 
We find that execution information and fine-tuning help LLMs maintain code correctness, and prompting with natural language often yields higher efficiency gains.
However, we broadly reconfirm findings that no existing method can improve efficiency without sacrificing functional correctness.
We hope \ds can serve as a testbed for program optimization, and we call for more efforts in advancing correctness-preserving program optimizations.

\section*{Acknowledgements}
We thank Aman Madaan for the helpful discussions in the early stage of the project.

\section*{Limitations}
Our benchmark establishes a solid foundation for rigorously evaluating the ability to generate efficient solutions, however, we also note that there are limitations to our work.

First, \ds has only included Python problems so far, but our \textsc{Judge0} evaluation platform and our benchmark curation recipe are fully reproducible and could be extended to other programming languages of interest.
Second, our benchmark currently focuses on competitive programming problems. It is also possible to extend our benchmark to other types of programming problems from more real-world software engineering scenarios.

Due to both limitations, our results may not be comprehensive enough to reflect the quality of model-generated programs on the full spectrum of all programming languages and problems. When using \ds in practice, we recommend the readers examine the model outputs, in addition to the quantitative results produced by our framework.


\bibliography{custom}

\begin{thebibliography}{36}
\expandafter\ifx\csname natexlab\endcsname\relax\def\natexlab#1{#1}\fi

\bibitem[{Baik et~al.(2024)Baik, Jeon, Hahn, Kim, Han, and Ko}]{baik2024codecomplex}
Seung-Yeop Baik, Mingi Jeon, Joonghyuk Hahn, Jungin Kim, Yo-Sub Han, and Sang-Ki Ko. 2024.
\newblock Codecomplex: A time-complexity dataset for bilingual source codes.
\newblock \emph{arXiv preprint arXiv:2401.08719}.

\bibitem[{Brown et~al.(2020)Brown, Mann, Ryder, Subbiah, Kaplan, Dhariwal, Neelakantan, Shyam, Sastry, Askell, Agarwal, Herbert-Voss, Krueger, Henighan, Child, Ramesh, Ziegler, Wu, Winter, Hesse, Chen, Sigler, Litwin, Gray, Chess, Clark, Berner, McCandlish, Radford, Sutskever, and Amodei}]{brown2020language}
Tom Brown, Benjamin Mann, Nick Ryder, Melanie Subbiah, Jared~D Kaplan, Prafulla Dhariwal, Arvind Neelakantan, Pranav Shyam, Girish Sastry, Amanda Askell, Sandhini Agarwal, Ariel Herbert-Voss, Gretchen Krueger, Tom Henighan, Rewon Child, Aditya Ramesh, Daniel Ziegler, Jeffrey Wu, Clemens Winter, Chris Hesse, Mark Chen, Eric Sigler, Mateusz Litwin, Scott Gray, Benjamin Chess, Jack Clark, Christopher Berner, Sam McCandlish, Alec Radford, Ilya Sutskever, and Dario Amodei. 2020.
\newblock \href {https://proceedings.neurips.cc/paper_files/paper/2020/file/1457c0d6bfcb4967418bfb8ac142f64a-Paper.pdf} {Language models are few-shot learners}.
\newblock In \emph{Advances in Neural Information Processing Systems}, volume~33, pages 1877--1901. Curran Associates, Inc.

\bibitem[{Bunel et~al.(2016)Bunel, Desmaison, Kumar, Torr, and Kohli}]{bunel2016superoptimprog}
Rudy Bunel, Alban Desmaison, M~Pawan Kumar, Philip~HS Torr, and Pushmeet Kohli. 2016.
\newblock Learning to superoptimize programs.
\newblock \emph{arXiv preprint arXiv:1611.01787}.

\bibitem[{Chen et~al.(2021)Chen, Tworek, Jun, Yuan, Pinto, Kaplan, Edwards, Burda, Joseph, Brockman et~al.}]{chen2021codexhumaneval}
Mark Chen, Jerry Tworek, Heewoo Jun, Qiming Yuan, Henrique Ponde de~Oliveira Pinto, Jared Kaplan, Harri Edwards, Yuri Burda, Nicholas Joseph, Greg Brockman, et~al. 2021.
\newblock Evaluating large language models trained on code.
\newblock \emph{arXiv preprint arXiv:2107.03374}.

\bibitem[{Coignion et~al.(2024)Coignion, Quinton, and Rouvoy}]{coignion2024performance}
Tristan Coignion, Cl{\'e}ment Quinton, and Romain Rouvoy. 2024.
\newblock A performance study of llm-generated code on leetcode.
\newblock In \emph{28th International Conference on Evaluation and Assessment in Software Engineering (EASE'24)}.

\bibitem[{Cummins et~al.(2023{\natexlab{a}})Cummins, Seeker, Grubisic, Elhoushi, Liang, Roziere, Gehring, Gloeckle, Hazelwood, Synnaeve et~al.}]{cummins2023large}
Chris Cummins, Volker Seeker, Dejan Grubisic, Mostafa Elhoushi, Youwei Liang, Baptiste Roziere, Jonas Gehring, Fabian Gloeckle, Kim Hazelwood, Gabriel Synnaeve, et~al. 2023{\natexlab{a}}.
\newblock Large language models for compiler optimization.
\newblock \emph{arXiv preprint arXiv:2309.07062}.

\bibitem[{Cummins et~al.(2023{\natexlab{b}})Cummins, Seeker, Grubisic, Elhoushi, Liang, Roziere, Gehring, Gloeckle, Hazelwood, Synnaeve et~al.}]{cummins2023metaoptim}
Chris Cummins, Volker Seeker, Dejan Grubisic, Mostafa Elhoushi, Youwei Liang, Baptiste Roziere, Jonas Gehring, Fabian Gloeckle, Kim Hazelwood, Gabriel Synnaeve, et~al. 2023{\natexlab{b}}.
\newblock Large language models for compiler optimization.
\newblock \emph{arXiv preprint arXiv:2309.07062}.

\bibitem[{Do{\v{s}}ilovi{\'c} and Mekterovi{\'c}(2020)}]{dovsilovic2020judge0}
Herman~Zvonimir Do{\v{s}}ilovi{\'c} and Igor Mekterovi{\'c}. 2020.
\newblock Robust and scalable online code execution system.
\newblock In \emph{2020 43rd International Convention on Information, Communication and Electronic Technology (MIPRO)}, pages 1627--1632. IEEE.

\bibitem[{Gao et~al.(2024)Gao, Gao, Gu, and Lyu}]{gao2024search}
Shuzheng Gao, Cuiyun Gao, Wenchao Gu, and Michael Lyu. 2024.
\newblock Search-based llms for code optimization.
\newblock \emph{arXiv preprint arXiv:2408.12159}.

\bibitem[{Gee et~al.(2024)Gee, Gritta, Lampouras, and Iacobacci}]{gee2024code}
Leonidas Gee, Milan Gritta, Gerasimos Lampouras, and Ignacio Iacobacci. 2024.
\newblock Code-optimise: Self-generated preference data for correctness and efficiency.
\newblock \emph{arXiv preprint arXiv:2406.12502}.

\bibitem[{Guo et~al.(2024)Guo, Zhu, Yang, Xie, Dong, Zhang, Chen, Bi, Wu, Li et~al.}]{guo2024deepseek}
Daya Guo, Qihao Zhu, Dejian Yang, Zhenda Xie, Kai Dong, Wentao Zhang, Guanting Chen, Xiao Bi, Y~Wu, YK~Li, et~al. 2024.
\newblock Deepseek-coder: When the large language model meets programming--the rise of code intelligence.
\newblock \emph{arXiv preprint arXiv:2401.14196}.

\bibitem[{Hu et~al.(2022)Hu, yelong shen, Wallis, Allen-Zhu, Li, Wang, Wang, and Chen}]{hu2022lora}
Edward~J Hu, yelong shen, Phillip Wallis, Zeyuan Allen-Zhu, Yuanzhi Li, Shean Wang, Lu~Wang, and Weizhu Chen. 2022.
\newblock \href {https://openreview.net/forum?id=nZeVKeeFYf9} {Lo{RA}: Low-rank adaptation of large language models}.
\newblock In \emph{International Conference on Learning Representations}.

\bibitem[{Huang et~al.(2024)Huang, Zhang, Qing, and Cui}]{huang2024effibench}
Dong Huang, Jie~M Zhang, Yuhao Qing, and Heming Cui. 2024.
\newblock Effibench: Benchmarking the efficiency of automatically generated code.
\newblock \emph{arXiv preprint arXiv:2402.02037}.

\bibitem[{Jeon et~al.(2023)Jeon, yeop Baik, Hahn, Han, and Ko}]{jeon2023deep}
Mingi Jeon, Seung yeop Baik, Joonghyuk Hahn, Yo-Sub Han, and Sang-Ki Ko. 2023.
\newblock \href {https://openreview.net/forum?id=9irBKvxsw9} {Deep learning-based source code complexity prediction}.

\bibitem[{Khan et~al.(2023)Khan, Bari, Do, Wang, Parvez, and Joty}]{khan2023xcodeeval}
Mohammad Abdullah~Matin Khan, M~Saiful Bari, Xuan~Long Do, Weishi Wang, Md~Rizwan Parvez, and Shafiq Joty. 2023.
\newblock xcodeeval: A large scale multilingual multitask benchmark for code understanding, generation, translation and retrieval.
\newblock \emph{arXiv preprint arXiv:2303.03004}.

\bibitem[{Kwon et~al.(2023)Kwon, Li, Zhuang, Sheng, Zheng, Yu, Gonzalez, Zhang, and Stoica}]{kwon2023efficient}
Woosuk Kwon, Zhuohan Li, Siyuan Zhuang, Ying Sheng, Lianmin Zheng, Cody~Hao Yu, Joseph~E. Gonzalez, Hao Zhang, and Ion Stoica. 2023.
\newblock Efficient memory management for large language model serving with pagedattention.
\newblock In \emph{Proceedings of the ACM SIGOPS 29th Symposium on Operating Systems Principles}.

\bibitem[{Li et~al.(2022)Li, Choi, Chung, Kushman, Schrittwieser, Leblond, Eccles, Keeling, Gimeno, Dal~Lago et~al.}]{li2022competition}
Yujia Li, David Choi, Junyoung Chung, Nate Kushman, Julian Schrittwieser, R{\'e}mi Leblond, Tom Eccles, James Keeling, Felix Gimeno, Agustin Dal~Lago, et~al. 2022.
\newblock Competition-level code generation with alphacode.
\newblock \emph{Science}, 378(6624):1092--1097.

\bibitem[{Liu et~al.(2024)Liu, Xie, Wang, Wei, Ding, and Zhang}]{liu2024evaluating}
Jiawei Liu, Songrun Xie, Junhao Wang, Yuxiang Wei, Yifeng Ding, and Lingming Zhang. 2024.
\newblock Evaluating language models for efficient code generation.
\newblock \emph{arXiv preprint arXiv:2408.06450}.

\bibitem[{Lozhkov et~al.(2024)Lozhkov, Li, Allal, Cassano, Lamy-Poirier, Tazi, Tang, Pykhtar, Liu, Wei et~al.}]{lozhkov2024starcoder}
Anton Lozhkov, Raymond Li, Loubna~Ben Allal, Federico Cassano, Joel Lamy-Poirier, Nouamane Tazi, Ao~Tang, Dmytro Pykhtar, Jiawei Liu, Yuxiang Wei, et~al. 2024.
\newblock Starcoder 2 and the stack v2: The next generation.
\newblock \emph{arXiv preprint arXiv:2402.19173}.

\bibitem[{Luo et~al.(2023)Luo, Xu, Zhao, Sun, Geng, Hu, Tao, Ma, Lin, and Jiang}]{luo2023wizardcoder}
Ziyang Luo, Can Xu, Pu~Zhao, Qingfeng Sun, Xiubo Geng, Wenxiang Hu, Chongyang Tao, Jing Ma, Qingwei Lin, and Daxin Jiang. 2023.
\newblock Wizardcoder: Empowering code large language models with evol-instruct.
\newblock \emph{arXiv preprint arXiv:2306.08568}.

\bibitem[{Madaan et~al.(2024)Madaan, Tandon, Gupta, Hallinan, Gao, Wiegreffe, Alon, Dziri, Prabhumoye, Yang et~al.}]{madaan2024self}
Aman Madaan, Niket Tandon, Prakhar Gupta, Skyler Hallinan, Luyu Gao, Sarah Wiegreffe, Uri Alon, Nouha Dziri, Shrimai Prabhumoye, Yiming Yang, et~al. 2024.
\newblock Self-refine: Iterative refinement with self-feedback.
\newblock \emph{Advances in Neural Information Processing Systems}, 36.

\bibitem[{Moudgalya et~al.(2023)Moudgalya, Ramakrishnan, Chemudupati, and Lu}]{moudgalya2023tasty}
Kaushik Moudgalya, Ankit Ramakrishnan, Vamsikrishna Chemudupati, and Xing~Han Lu. 2023.
\newblock Tasty: A transformer based approach to space and time complexity.
\newblock \emph{arXiv preprint arXiv:2305.05379}.

\bibitem[{Niu et~al.(2024)Niu, Zhang, Li, Luo, and Ng}]{niu2024evaluating}
Changan Niu, Ting Zhang, Chuanyi Li, Bin Luo, and Vincent Ng. 2024.
\newblock On evaluating the efficiency of source code generated by llms.
\newblock \emph{arXiv preprint arXiv:2404.06041}.

\bibitem[{Ouyang et~al.(2022)Ouyang, Wu, Jiang, Almeida, Wainwright, Mishkin, Zhang, Agarwal, Slama, Ray et~al.}]{ouyang2022training}
Long Ouyang, Jeffrey Wu, Xu~Jiang, Diogo Almeida, Carroll Wainwright, Pamela Mishkin, Chong Zhang, Sandhini Agarwal, Katarina Slama, Alex Ray, et~al. 2022.
\newblock Training language models to follow instructions with human feedback.
\newblock \emph{Advances in neural information processing systems}, 35:27730--27744.

\bibitem[{Puri et~al.(2021)Puri, Kung, Janssen, Zhang, Domeniconi, Zolotov, Dolby, Chen, Choudhury, Decker et~al.}]{puri2021codenet}
Ruchir Puri, David~S Kung, Geert Janssen, Wei Zhang, Giacomo Domeniconi, Vladimir Zolotov, Julian Dolby, Jie Chen, Mihir Choudhury, Lindsey Decker, et~al. 2021.
\newblock Codenet: A large-scale ai for code dataset for learning a diversity of coding tasks.
\newblock \emph{arXiv preprint arXiv:2105.12655}.

\bibitem[{Ridnik et~al.(2024)Ridnik, Kredo, and Friedman}]{ridnik2024alphacodium}
Tal Ridnik, Dedy Kredo, and Itamar Friedman. 2024.
\newblock Code generation with alphacodium: From prompt engineering to flow engineering.
\newblock \emph{arXiv preprint arXiv:2401.08500}.

\bibitem[{Roziere et~al.(2023)Roziere, Gehring, Gloeckle, Sootla, Gat, Tan, Adi, Liu, Remez, Rapin et~al.}]{roziere2023code}
Baptiste Roziere, Jonas Gehring, Fabian Gloeckle, Sten Sootla, Itai Gat, Xiaoqing~Ellen Tan, Yossi Adi, Jingyu Liu, Tal Remez, J{\'e}r{\'e}my Rapin, et~al. 2023.
\newblock Code llama: Open foundation models for code.
\newblock \emph{arXiv preprint arXiv:2308.12950}.

\bibitem[{Shi and Zhang(2020)}]{shi2020symbsuperoptim}
Hui Shi and Yang Zhang. 2020.
\newblock Deep symbolic superoptimization without human knowledge.
\newblock \emph{ICLR 2020}.

\bibitem[{Shinn et~al.(2024)Shinn, Cassano, Gopinath, Narasimhan, and Yao}]{shinn2024reflexion}
Noah Shinn, Federico Cassano, Ashwin Gopinath, Karthik Narasimhan, and Shunyu Yao. 2024.
\newblock Reflexion: Language agents with verbal reinforcement learning.
\newblock \emph{Advances in Neural Information Processing Systems}, 36.

\bibitem[{Shypula et~al.(2021)Shypula, Yin, Lacomis, Goues, Schwartz, and Neubig}]{shypula2021learning}
Alex Shypula, Pengcheng Yin, Jeremy Lacomis, Claire~Le Goues, Edward Schwartz, and Graham Neubig. 2021.
\newblock Learning to superoptimize real-world programs.
\newblock \emph{arXiv preprint arXiv:2109.13498}.

\bibitem[{Shypula et~al.(2024)Shypula, Madaan, Zeng, Alon, Gardner, Yang, Hashemi, Neubig, Ranganathan, Bastani, and Yazdanbakhsh}]{shypula2024learning}
Alexander~G Shypula, Aman Madaan, Yimeng Zeng, Uri Alon, Jacob~R. Gardner, Yiming Yang, Milad Hashemi, Graham Neubig, Parthasarathy Ranganathan, Osbert Bastani, and Amir Yazdanbakhsh. 2024.
\newblock \href {https://openreview.net/forum?id=ix7rLVHXyY} {Learning performance-improving code edits}.
\newblock In \emph{The Twelfth International Conference on Learning Representations}.

\bibitem[{Sikka et~al.(2020)Sikka, Satya, Kumar, Uppal, Shah, and Zimmermann}]{sikka2020corcod}
Jagriti Sikka, Kushal Satya, Yaman Kumar, Shagun Uppal, Rajiv~Ratn Shah, and Roger Zimmermann. 2020.
\newblock Learning based methods for code runtime complexity prediction.
\newblock In \emph{Advances in Information Retrieval: 42nd European Conference on IR Research, ECIR 2020, Lisbon, Portugal, April 14--17, 2020, Proceedings, Part I 42}, pages 313--325. Springer.

\bibitem[{Singhal et~al.(2024)Singhal, Aggarwal, Awasthi, Natarajan, and Kanade}]{singhal2024nofuneval}
Manav Singhal, Tushar Aggarwal, Abhijeet Awasthi, Nagarajan Natarajan, and Aditya Kanade. 2024.
\newblock Nofuneval: Funny how code lms falter on requirements beyond functional correctness.
\newblock \emph{arXiv preprint arXiv:2401.15963}.

\bibitem[{Team et~al.(2024)Team, Mesnard, Hardin, Dadashi, Bhupatiraju, Pathak, Sifre, Rivi{\`e}re, Kale, Love et~al.}]{team2024gemma}
Gemma Team, Thomas Mesnard, Cassidy Hardin, Robert Dadashi, Surya Bhupatiraju, Shreya Pathak, Laurent Sifre, Morgane Rivi{\`e}re, Mihir~Sanjay Kale, Juliette Love, et~al. 2024.
\newblock Gemma: Open models based on gemini research and technology.
\newblock \emph{arXiv preprint arXiv:2403.08295}.

\bibitem[{Wei et~al.(2022)Wei, Bosma, Zhao, Guu, Yu, Lester, Du, Dai, and Le}]{wei2022finetuned}
Jason Wei, Maarten Bosma, Vincent Zhao, Kelvin Guu, Adams~Wei Yu, Brian Lester, Nan Du, Andrew~M. Dai, and Quoc~V Le. 2022.
\newblock \href {https://openreview.net/forum?id=gEZrGCozdqR} {Finetuned language models are zero-shot learners}.
\newblock In \emph{International Conference on Learning Representations}.

\bibitem[{Ye et~al.(2024)Ye, Ma, Wu, Zhang, Ji, and Wang}]{ye2024iterative}
Tong Ye, Tengfei Ma, Lingfei Wu, Xuhong Zhang, Shouling Ji, and Wenhai Wang. 2024.
\newblock Iterative or innovative? a problem-oriented perspective for code optimization.
\newblock \emph{arXiv preprint arXiv:2406.11935}.

\end{thebibliography}

\clearpage
\appendix

\section{Iterative Refinement Analysis}
\label{appendix:refinement_per_opt}
Following \citet{shypula2024learning}, we also evaluate runtime and memory efficiency as a percentage of pairs where the generated code $p_{out}$ is faster/uses lesser memory than $p_{in}$, in addition to the speedup and memory reduction metrics described in \S\ref{sub:3.3.2:history-based editing}. 

In \autoref{tab:refine-optim-perc}, these percentage optimized metrics clearly indicate that exec-refine is the most consistent iterative refinement method as it achieves the best pass@1, \% runtime optimized and \% memory optimized across all models. Contrasting these results to Speedup and Memory Reduction in \autoref{tab:refine-results}, we note that while natural language feedback (in self-refine and NL+Exec refine) aids in significantly improving the runtime and memory usage for a few cases (while breaking test cases for others), exec-refine improves runtime and memory in more cases albeit to a smaller degree.

\begin{table*}[h!]
    \vspace{-3mm}
      \centering
      \begin{tabular}{ll|ccc}
        \toprule
        \multicolumn{1}{c}{\multirow{2}{*}{\textbf{Model}}} & \multicolumn{1}{c|}{\multirow{2}{*}{\textbf{Setting}}} & \multicolumn{3}{c}{\textbf{History-based Editing}}  \\
        {} & {} & {pass@1} & {\% Runtime Opt} & {\% Mem. Opt.} \\
        \midrule
        \multirow{3}{*}{StarCoder2} & {self-refine} & {26.7} & {22.5} & {27.0} \\
        {} & {exec-refine} & {\bf 39.5} & {\bf 25.5} & {\bf 30.3} \\
        {} & {nl+exec refine} & {26.1} & {22.6} & {27.0} \\
        \midrule
        \multirow{3}{*}{CodeGemma} & {self-refine} & {15.1} & {18.2} & {29.3} \\
        {} & {exec-refine} & {\bf 33.2} & {\bf 25.5} & {\bf 34.5} \\
        {} & {nl+exec refine} & {29.8} & {21.0} & {31.3} \\
        \midrule
        \multirow{3}{*}{WizardCoder} & {self-refine} & {8.5} & {16.1} & {24.4} \\
        {} & {exec-refine} & {\bf 20.9} & {\bf 20.0} & {\bf 27.2} \\
        {} & {nl+exec refine} & {18.3} & {17.1} & {26.9} \\
        \midrule
        \multirow{3}{*}{CodeLLaMa} & {self-refine} & {15.8} & {15.6} & {24.5} \\
        {} & {exec-refine} & {\bf 54.6} & {\bf 27.4} & {\bf 39.6} \\
        {} & {nl+exec refine} & {16.2} & {16.8} & {23.2} \\
        \midrule
        \multirow{3}{*}{DeepseekCoder} & {self-refine} & {13.6} & {26.8} & {35.8} \\
        {} & {exec-refine} & {\bf 27.4} & {\bf 31.4} & {\bf 40.0} \\
        {} & {nl+exec refine} & {19.6} & {29.7} & {36.5} \\
        \midrule
        \multirow{3}{*}{GPT-4o} & {self-refine} & {47.8} & {42.5} & {42.3} \\
        {} & {exec-refine} & {\bf60.8} & {\bf 48.8} & {\bf 51.1} \\
        {} & {nl+exec refine} & {58.8} & {48.6} & {48.1} \\
        \bottomrule
      \end{tabular}
      \caption{Comparing iterative refinement approaches on \% Optimization metrics. Exec-Refine is the best performing approach across all models and metrics.}
      \label{tab:refine-optim-perc}
    \vspace{-2mm}
\end{table*}

\section{Implementation Details}
\label{app:implementation}
\paragraph{Generation} We use a maximum length of 1024 tokens, and a temperature of $t=0.4$. We provide two examples for all in-context few-shot experiments

\paragraph{Hardware} We run all our experiments on a mix of A6000 and L40 GPUs. Specifically for the prompting iterative prompting and in-context learning approaches, we use 1-2 GPUs of the type and utilize the vLLM library \cite{kwon2023efficient} primarily for generating programs efficiently. We perform finetuning on 4 A6000 GPUs.

For \textsc{Judge0} evaluation virtual hardware setup,  we use an m7i.large EC2 instance which has 2 vCPU cores of the 4th Generation Intel Xeon Sapphire Rapids, with 8GB of RAM.

\paragraph{Fine-tuning} We adopt parameter-efficient fine-tuning with LoRA \citep{hu2022lora} due to resource limitations and implement this using the HuggingFace Transformers library \footnote{https://github.com/huggingface/transformers}. We optimize the finetuning using Deepspeed \footnote{https://github.com/microsoft/DeepSpeed} ZeRo stage 2 with a LoRA rank of 8, alpha of 16 and a per-device batch size of 2. We use the AdamW optimizer, with a learning rate of 1e-3 and a warmup of 100 steps. We train the models on the history-based editing task for a single epoch and fine-tune models for 10 epochs on the NL-instructed generation task.

\section{Variance Analysis}
We measured the variance for two models (DeepseekCoder and CodeLLaMa) by sampling generations three times and evaluating each generation independently on our Judge setup. The results are included below with 95\% Confidence Intervals are shown in \autoref{tab:variance-history-nl-generation-results}.

Pass@1 has no variance across runs in both paradigms due to the low temperature; runtime efficiency metrics (speedup and runtime \%) both only have about ~1\% Confidence Interval (CI). Memory has close to no variance in the history-based editing setting but shows ~5\% CI in NL-instructed generation.
\begin{table*}[h!]
    \vspace{-3mm}
    \centering
    \resizebox{0.96\textwidth}{!}{
        \begin{tabular}{l|ccc|ccc}
            \toprule
            \multicolumn{1}{c|}{\textbf{Model}} & \multicolumn{3}{c|}{\textbf{History-based Editing}} & \multicolumn{3}{c}{\textbf{NL-Instructed Generation}} \\
            {} & {Pass@1} & {Speedup} & {Memory Reduction} & {Pass@1} & {Runtime (\%)} & {Memory (\%)} \\
            \midrule
            DeepseekCoder & 32.87 $\pm$ 0.00 & 2.112 $\pm$ 0.022 & 1.097 $\pm$ 0.000166 & 18.75 $\pm$ 0.00 & 57.589 $\pm$ 1.012 & 67.796 $\pm$ 3.106 \\
            CodeLLaMa & 57.50 $\pm$ 0.00 & 1.456 $\pm$ 0.0028 & 1.114 $\pm$ 0.00173 & 8.30 $\pm$ 0.00 & 55.663 $\pm$ 0.720 & 61.029 $\pm$ 3.818 \\
            \bottomrule
        \end{tabular}
    }
    \caption{Variance of results for History-based Editing and NL-Instructed Generation}
    \label{tab:variance-history-nl-generation-results}
    \vspace{-2mm}
\end{table*}

\begin{table*}[ht]
\small
\centering
\resizebox{0.96\textwidth}{!}{
  \begin{tabular}{ll|ccc|ccc}
    \toprule
    \multicolumn{1}{c}{\multirow{2}{*}{\textbf{Model}}} & \multicolumn{1}{c|}{\multirow{2}{*}{\textbf{Size}}} & \multicolumn{3}{c|}{\textbf{History-based Editing}} & \multicolumn{3}{c}{\textbf{NL-instructed Generation}} \\
    {} & {} & {Pass@1} & {Speedup} & {Memory Reduction} & {Pass@1} & {Runtime (\%)} & {Memory (\%)} \\
    \midrule
    \multirow{4}{*}{CodeLLaMa} 
    & {~~7B}  & {58.4} & {1.04} & {1.00} & {~~2.1} & {39.73} & {26.93} \\
    {} & {13B} & {57.5} & {1.44} & {1.11} & {~~8.3} & {45.30} & {74.18} \\
    {} & {34B} & {51.9} & {1.31} & {1.13} & {~~8.3} & {43.23} & {70.16} \\
    {} & {70B} & {55.3} & {1.73} & {1.09} & {12.5} & {39.15} & {64.89} \\
    \midrule
    \multirow{3}{*}{DeepseekCoder} 
    & {1.3B} & {44.2} & {1.36} & {1.13} & {~~8.3} & {46.71} & {65.62} \\
    {} & {6.7B} & {34.8} & {1.47} & {1.16} & {29.2} & {33.48} & {76.19} \\
    {} & {~33B} & {49.1} & {1.51} & {1.09} & {37.5} & {43.91} & {75.61} \\
    \bottomrule
  \end{tabular}
}
\caption{Results for History-based Editing and NL-Instructed Generation across different model sizes.}
\vspace{-1em}
\label{tab:scaling-results}
\end{table*}

\section{Prompt Details}
\label{app:prompt-details}

We illustrate and detail all of the prompts used for the experiments in Figures \ref{fig:prompt-inst}-\ref{fig:prompt-traj-cond}. 

\begin{figure*}[h]
\centering
\begin{minipage}{\textwidth}
\begin{mdframed}[
    backgroundcolor=black!10,
    outerlinewidth=2pt,
    outerlinecolor=blue,
    roundcorner=10pt,
    innertopmargin=10pt,
    innerbottommargin=10pt,
    innerrightmargin=15pt,
    innerleftmargin=15pt
]
\begin{verbatim}
Optimize the python program below to be functionally equivalent 
but run faster and use less memory.

Here are a few examples:
            
### Program:
{slow_code_example} 

### Optimized (Runtime and Space) version of Program above:
{fast_code_example}

### Program:
[src_code]

### Optimized (Runtime and Space) version of Program above:
\end{verbatim}
\end{mdframed}
\end{minipage}
\caption{Prompt for Instruction prompting $I_{eff}$ along with in-context examples}
\label{fig:prompt-inst}
\end{figure*}

\begin{figure*}[h]
\centering
\begin{minipage}{\textwidth}
\begin{mdframed}[
    backgroundcolor=black!10,
    outerlinewidth=2pt,
    outerlinecolor=blue,
    roundcorner=10pt,
    innertopmargin=10pt,
    innerbottommargin=10pt,
    innerrightmargin=15pt,
    innerleftmargin=15pt
]
\begin{verbatim}
Your solution was functionally {CORRECT/INCORRECT}

Here are the run time and memory stats of your code for each test case
-- Stats for test case 0 --
Correct: {PASSED/FAILED}
Run time: 0.009 s
Memory: 3352.0 KB
-- Stats for test case 1 --
Correct: {PASSED/FAILED}
Run time: 0.009 s
Memory: 3316.0 KB
\end{verbatim}
\end{mdframed}
\end{minipage}
\label{fig:exec-feedback}
\caption{Example of Execution Feedback on public test cases used for Exec-Refine and Execution Conditioned Fine-tuning}
\end{figure*}

\begin{figure*}[h]
\centering
\begin{minipage}{\textwidth}
\begin{mdframed}[
    backgroundcolor=black!10,
    outerlinewidth=2pt,
    outerlinecolor=blue,
    roundcorner=10pt,
    innertopmargin=10pt,
    innerbottommargin=10pt,
    innerrightmargin=15pt,
    innerleftmargin=15pt
]

\begin{verbatim}
Write a python code which is correct and efficient in terms of runtime and memory 
usage for the following problem description.

##Problem Name:
{problem_name}

##Problem Description:
{In detail description of the task}

## Sample Inputs:
{input_test_cases}

##Sample Outputs:
{Expected Output}
\end{verbatim}
\end{mdframed}
\end{minipage}
\caption{Prompt for NL-instructed generation $I_{gen}$}
\end{figure*}
\begin{figure*}[h]
\centering
\begin{minipage}{\textwidth}
\begin{mdframed}[
    backgroundcolor=black!10,
    outerlinewidth=2pt,
    outerlinecolor=blue,
    roundcorner=10pt,
    innertopmargin=10pt,
    innerbottommargin=10pt,
    innerrightmargin=15pt,
    innerleftmargin=15pt
]
\begin{verbatim}
Give feedback in english for why the code solution below is incorrect 
or inefficient and how the program can be fixed.

## Candidate solution:
{most recent code attempt}

## Feedback for incorrectness/inefficiency and how it can be improved:
\end{verbatim}
\end{mdframed}
\end{minipage}
\caption{Prompt used for NL-reasoning to get feedback}
\end{figure*}

\begin{figure*}[h]
\centering
\begin{minipage}{\textwidth}
\begin{mdframed}[
    backgroundcolor=black!10,
    outerlinewidth=2pt,
    outerlinecolor=blue,
    roundcorner=10pt,
    innertopmargin=10pt,
    innerbottommargin=10pt,
    innerrightmargin=15pt,
    innerleftmargin=15pt
]
\begin{verbatim}
Refine the given incorrect or sub-optimal code solution based on the feedback 
specified below.

### Candidate solution: 
{previous_code_attempt}

### Feedback for incorrectnes/inefficiency and how it can be improved:
{self-feedback / execution-feedback / NL+Exec-Refine}

### Optimized/Corrected solution based on feedback:
\end{verbatim}
\end{mdframed}
\end{minipage}
\caption{Prompt used for refining code in Self-Refine, Exec-Refine and NL+Exec-Refine}
\end{figure*}

\begin{figure*}[h]
\centering
\begin{minipage}{\textwidth}
\begin{mdframed}[
    backgroundcolor=black!10,
    outerlinewidth=2pt,
    outerlinecolor=blue,
    roundcorner=10pt,
    innertopmargin=10pt,
    innerbottommargin=10pt,
    innerrightmargin=15pt,
    innerleftmargin=15pt
]
\begin{verbatim}
Based on the execution results, reflect on why the code solution 
below was incorrect or inefficient and how the program can be fixed.

{generated code solution}

## Execution Results:
{Execution Feedback for the previous attempt}

### Reflection on incorrectnes/inefficiency and how it can be improved:
\end{verbatim}
\end{mdframed}
\end{minipage}
\caption{Prompt used for NL+Exec-Refine to reflect on Execution results}
\end{figure*}

\begin{figure*}[h]
\centering
\begin{minipage}{\textwidth}
\begin{mdframed}[
    backgroundcolor=black!10,
    outerlinewidth=2pt,
    outerlinecolor=blue,
    roundcorner=10pt,
    innertopmargin=10pt,
    innerbottommargin=10pt,
    innerrightmargin=15pt,
    innerleftmargin=15pt
]

\begin{verbatim}
##1 iteration program:
{slowest program in trajectory}

##2 iteration program:
{33 percentile fastest program in trajectory}

##3 iteration program:
{66 percentile fastest program in trajectory}

### Final iteration program:
\end{verbatim}
\end{mdframed}
\end{minipage}
\caption{Format of Trajectory-Conditioned Fine-tuning data}
\label{fig:prompt-traj-cond}
\end{figure*}

\end{document}